\documentclass[journal]{IEEEtran}
\usepackage{amsmath,amsfonts}
\usepackage{algorithmic}
\usepackage{algorithm}
\usepackage{array}
\usepackage[caption=false,font=normalsize,labelfont=sf,textfont=sf]{subfig}
\usepackage{textcomp}
\usepackage{stfloats}
\usepackage{url}
\usepackage{verbatim}
\usepackage{graphicx}
\usepackage{cite}
\usepackage{booktabs}
\usepackage{multirow}
\usepackage{verbatim}
\begin{document}

\title{A Deep Hierarchical Feature Sparse Framework for Occluded Person Re-Identification}

\author{Yihu Song} 
 
\markboth{Journal of \LaTeX\ Class Files,~Vol.~14, No.~8, August~2021}%
{Shell \MakeLowercase{\textit{et al.}}: A Sample Article Using IEEEtran.cls for IEEE Journals}


\maketitle

\begin{abstract}
Most existing methods tackle the problem of occluded person re-identification (ReID) by utilizing auxiliary models, resulting in a complicated and inefficient ReID framework that is unacceptable for real-time applications. In this work, a speed-up person ReID framework named SUReID is proposed to mitigate occlusion interference while speeding up inference. The SUReID consists of three key components: hierarchical token sparsification (HTS) strategy, non-parametric feature alignment knowledge distillation (NPKD), and noise occlusion data augmentation (NODA). The HTS strategy works by pruning the redundant tokens in the vision transformer to achieve highly effective self-attention computation and eliminate interference from occlusions or background noise. However, the pruned tokens may contain human part features that contaminate the feature representation and degrade the performance. To solve this problem, the NPKD is employed to supervise the HTS strategy, retaining more discriminative tokens and discarding meaningless ones. Furthermore, the NODA is designed to introduce more noisy samples, which further trains the ability of the HTS to disentangle different tokens. Experimental results show that the SUReID achieves superior performance with surprisingly fast inference.
\end{abstract}

\begin{IEEEkeywords}
occluded person re-identification, efficient vision transformer, knowledge distillation.
\end{IEEEkeywords}

\section{Introduction}
\IEEEPARstart{P}{erson} ReID task aims to identify and retrieve target pedestrians captured by non-overlapping cameras. It is an important topic in the field of computer vision with a wide range of practical applications, including video surveillance, security, and smart cities \cite{ref1,ref2}. With the rapid development of deep learning, holistic person ReID is making great progress, and various methods have been proposed \cite{ref3,ref4,ref5,ref6}. However, pedestrians are often occluded by various obstacles (e.g., cars, trees, walls, and other people), making it difficult for holistic person re-identification methods to perform well in such scenarios. To address this issue, occluded person ReID has attracted the attention of researchers, and some feasible solutions have been proposed \cite{ref7,ref8,ref9,ref10}.

\begin{figure}[htbp]
\centering
\includegraphics[width=0.3\textwidth]{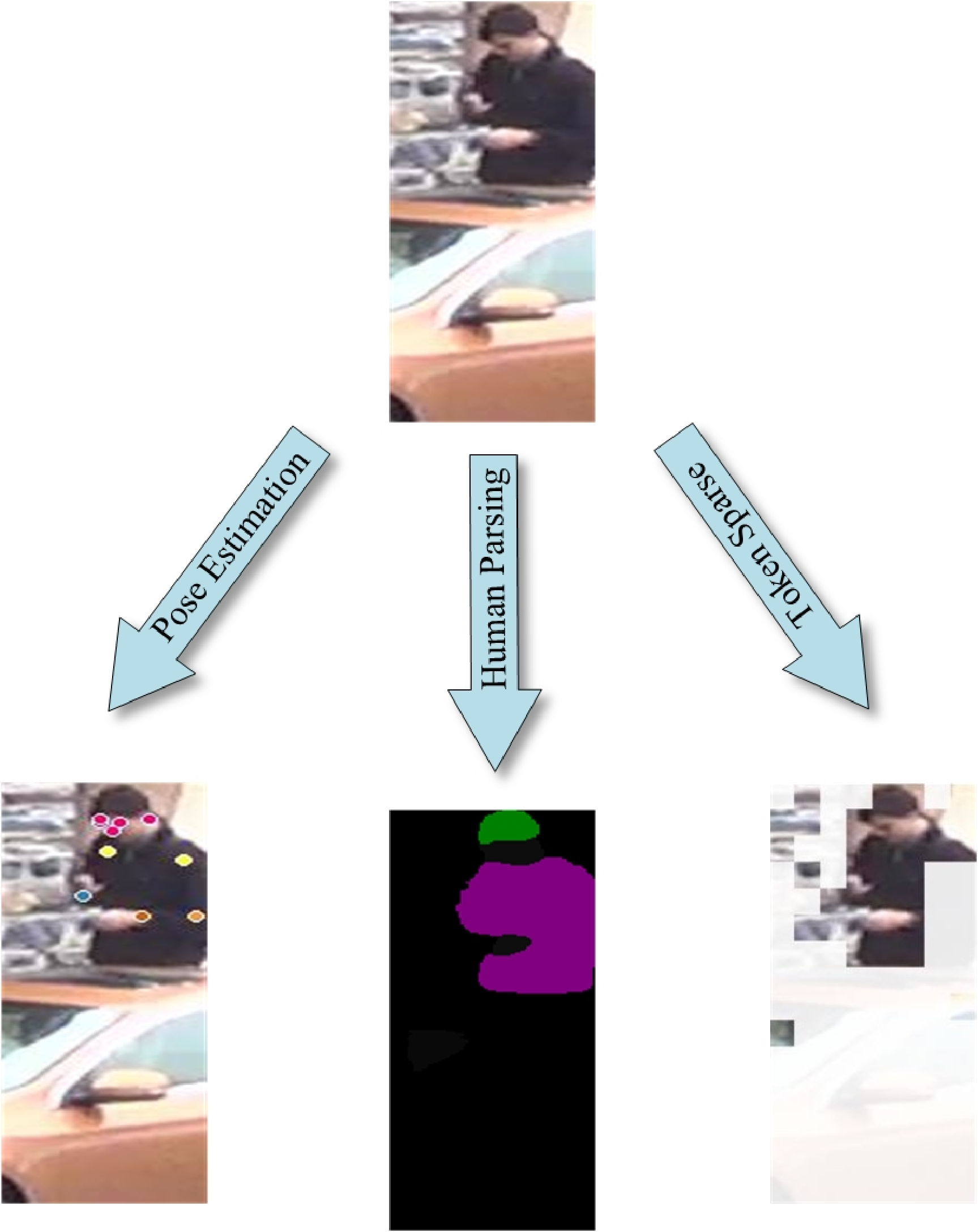}
\caption{Illustration of the different methods to tackle occluded ReID task.}\label{fig1}
\end{figure} 

Comparing with the holistic ReID task, occluded person ReID is more challenging since occlusions inevitably lead to incomplete body information and spatial misalignment. In particular, the different occlusions introduce intra-class variations that cause more errors in image matching. Furthermore, the presence of some occlusions with similar appearances can degrade the learned person image representation. To resolve the problem intuitively, a discriminative feature can be learned from unobstructed regions. As illustrated in Fig.\ref{fig1}, many current methods aim to detect non-occluded body parts and align visible body part features by utilizing external cues, such as human parsing or pose estimation models \cite{ref11,ref12,ref13}. With the help of these external cues, these methods can effectively avoid interference caused by non-pedestrian obstructions, resulting in more accurate matching of person images. However, these methods often do not prioritize target pedestrians when they are obstructed by other irrelevant pedestrians \cite{ref14}. Moreover, involving external models in the inference phase leads to significant computational demands, making them unsuitable for real-time applications \cite{ref15}. 

To tackle the aforementioned issues, this paper presents SUReID, an efficient and robust ReID framework that eliminates occlusions through token sparsification strategy. Thanks to the inherent nature of self-attention operations in transformer, the acceleration of unstructured token sets produced by the HTS strategy becomes readily achievable through parallel computing \cite{ref16}. As illustrated in Fig.\ref{fig1}, the masked patches are uninformative and will be discarded. The final prediction is made by considering only a subset of the most informative tokens, which proves to be adequate for achieving precise image recognition. In addition, the SUReID proposes NPKD to supervise the kept token contains more valuable information. Actually, the existing knowledge distillation methods are unsuitable towards person ReID task since they typically distill the logits value that are not used in the calculation of testing phase \cite{ref17,ref18,ref19}. More than just logits knowledge distillation, the NPKD also takes into account the feature-based knowledge distillation of class token \cite{ref20,ref21,ref22, ref23}. As the feature dimensions may mismatch between the teacher model and student model, the NPKD employs a simple interpolation technique to align them. Additionally, this paper proposes NODA to introduce occluded samples which are commonly found in real-life scenarios. Since the occluded samples are irrelevant to the training dataset, this data augmentation strategy usually contaminates the learned features. Nevertheless, owing to the HTS strategy, the noise occlusion information can assist the ReID model to concentrate on more valuable human features. In the inference phase, the SUReID does not require any external cues to supervise the ReID model in locating the discriminative part. Additionally, the ReID model only calculates retained tokens, demonstrating robustness against occlusions and background noise, while exhibiting high efficiency in processing images of individuals.

The main contributions of this paper can be summarized as follows. First, the HTS strategy is developed to overcome the occlusion problem while speeding up the inference. Second, the NPDK is proposed to distill prior knowledge from a pre-trained ReID model to improve the feature representation capability of kept tokens. Third, the NODA provides more noisy samples, which further trains the ability of the SUReID for disentangling the discriminative human body parts. Based on the above investigations, the SUReID outperforms other methods in the occluded ReID task with higher speed of inference.

The rest of this article is organized as follows. First, some related works are reviewed and discussed in section \ref{sec2}. Then, a detailed illustration of the proposed method, including the HTS strategy, NPKD and NODA, is given in section \ref{sec3}. The experimental results and analyses are presented in section \ref{sec4}. Finally, section \ref{sec5} concludes this paper and section \ref{sec6} shows the limitation of this method and illustrates the future work.

\section{Related work}\label{sec2}

This section briefly overviews some works related to occluded person ReID methods, efficient transformer, and knowledge distillation. 

\subsection{Occluded person ReID}\label{subsec2}

Given occluded probe images, occluded person ReID aims to identify the same person using their full-body appearance captured by different cameras. Recent methods address this task by utilizing external cues, such as incorporating pose estimation or human parsing as assistance. Miao et al. introduce pose-guided feature alignment (PGFA) to disentangle discriminative human parts from occlusion features by utilizing pose landmarks. Gao et al. present a pose-guided visible part matching (PVPM) method that simultaneously learns the discriminative features with pose-guided attention and graph matching strategy in an end-to-end framework. Wang et al. propose a robust feature alignment approach by jointly optimizing high-order relations using graph convolution layers and leveraging human-topology information through key-point estimation. The above methods align visible body feature precisely according to the guidance of additional pose estimation model. However, they are impractical for deployment due to the enormous computational burden brought by the external models. In contrast to the aforementioned strict alignment-based approaches, the proposed SUReID framework employs a pretrained teacher model to assist in locating the discriminative human part. More importantly, the teacher model will be discarded in favour of a lightweight student model for calculation during inference \cite{ref24}. The other popular methods employ attention mechanisms to tackle occluded person ReID tasks. The self-attention mechanism shows strong performance as it can globally model the relationships between feature representations of different semantic components \cite{ref25}. He et al. investigate a pure transformer framework named TransReID for the object ReID task. The results demonstrate the robustness of the self-attention mechanism. Wang et al. develop a pose-guided feature disentangling (PFD) framework to train both a pure transformer network and a pose estimation model with learnable parameters.Despite its superior performance, the computation of quadratic times dot-product in the self-attention mechanism and the addition of the attitude guidance model slows its running speed. In contrast to transformer-based methods, SUReID is a pure transformer network that uses HTS strategy to prune redundant tokens, making it an efficient inference model.

\subsection{Efficient Transformer}\label{subsec2}

Vision transformer (ViT) has demonstrated remarkable performance in various vision tasks \cite{ref26}. However, the self-attention mechanism in ViT suffers relatively intensive computational cost due to the quadratic number of interactions between tokens. To address this issue, several research studies have proposed methods to build a more efficient transformer. Recent researches have discovered that the self-attention mechanism in vision transformers exhibits sparsity. Accordingly, some works propose to prune tokens based on importance score in transformer. Rong et al. present DynamicViT (dynamic vision transformer), which designs a lightweight prediction module to estimate the importance score of each token to prune redundant tokens. Meng et al. introduce AdaViT (adaptive vision transformer) that learns to derive usage policies on which patches, self-attention heads and transformer blocks to keep throughput the transformer backbone \cite{ref27}. The above two methods perform token pruning by extra light-weight prediction modules. The prediction modules record informative tokens information in training phase, and discard them during inference. The other approach leverages the class token attention to keep attentive tokens and prune informative tokens. Liang et al. propose EViT (expending vision transformer), it defines the attentive tokens as image tokens with the largest attention value from the class token, and fuses the information from less informative tokens to a new token \cite{ref28}. Yin et al. design A-ViT (adaptive tokens vision transformer), an adaptive token pruning mechanism is investigated based on class token attention, which dynamically adjusts the calculation cost of images with different complexity \cite{ref29}. Following DyViT, the HTS strategy also integrates an additional lightweight prediction module to determine which tokens to discard. Furthermore, inspired by the class token attention-based methods, the SUReID proposes a class token attention reweight module on the top of the student model to enhance the feature representation capability of attentive tokens.

\begin{figure*}[htbp]
\centering
\includegraphics[width=0.55\textwidth]{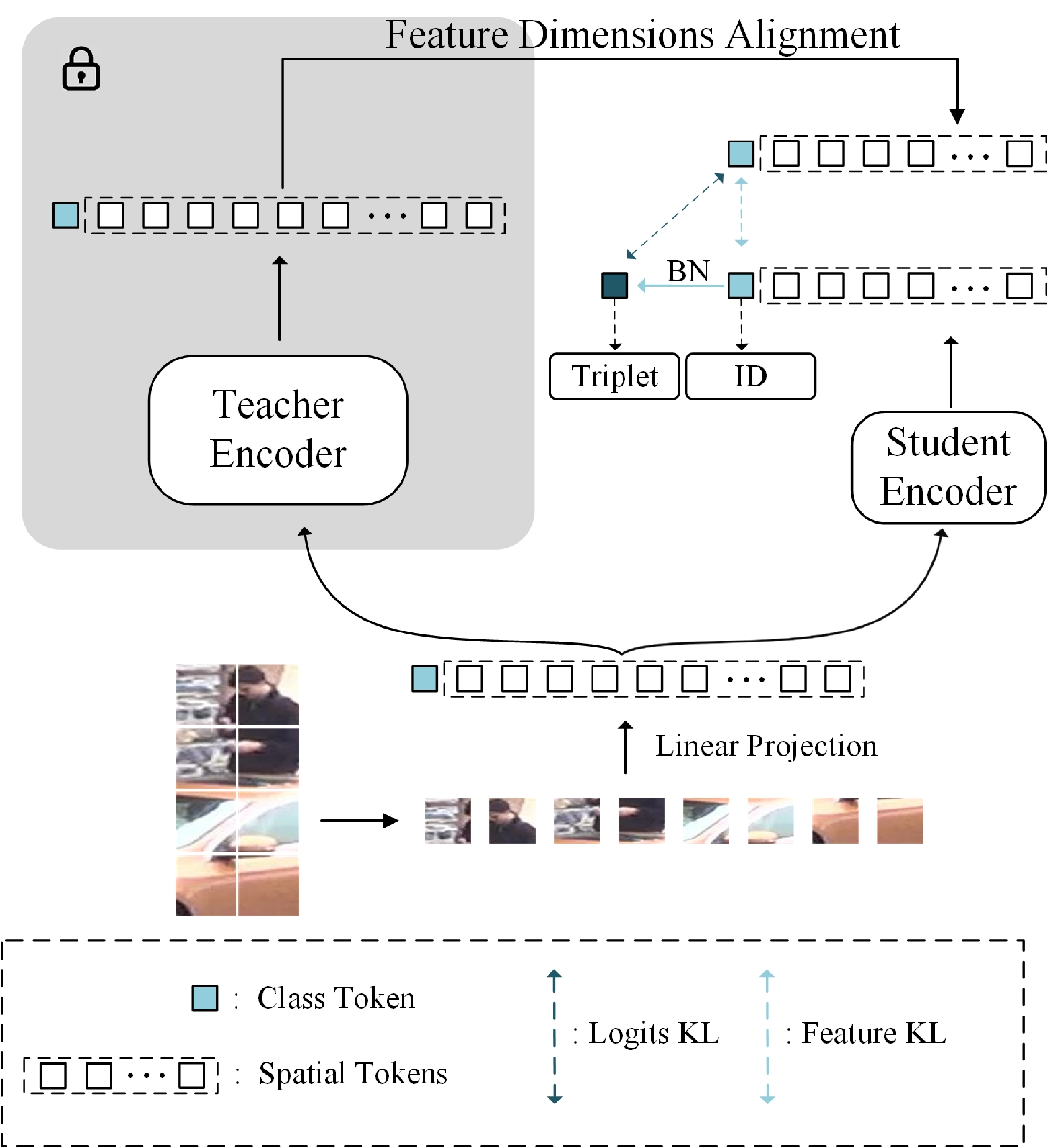}
\caption{The pipeline of SUReID framework.}\label{fig2}
\end{figure*}

\subsection{Knowledge distillation}\label{subsec2}

Knowledge distillation (KD) has been developed as an efficient means of compressing and accelerating models. The transferred knowledge is initially considered as the conditional distribution of outputs given input samples. KD methods can be roughly divided into two groups: response-based and feature-based methods. Response-based methods are employed in classification tasks to distill the final logits from teacher to student through minimizing the Kullback-Leibler (KL) divergence. However, the response-based KD overlooks significant information of the intermediate features, which are found to be crucial for representation learning. For example, the person ReID task only computes the similarity of the output feature embeddings during testing phase and discards the distilled logits information. As a result, response-based logits KD may not be suitable for ReID tasks. Feature-based KD aligns the distributions between the teacher and student in the embedding space. Romero et al. introduce FitNets, which transfer knowledge by utilizing both final and intermediate outputs. The approach incorporates a regressor on the intermediate layers to align teacher's and student's outputs of varying sizes. Additionally, attention maps, neuron selectivity patterns, paraphrasers, and route constraints inspired by FitNets are proposed to further utilize feature-based knowledge \cite{ref30,ref31,ref32, ref33}. The proposed SUReID in this paper distills prior knowledge from a pretrained teacher model to student model by simultaneously leveraging the feature-based KD with a larger weight and the response-based KD with a smaller weight.

\section{Proposed method}\label{sec3}

This section briefly introduces the proposed SUReID, the overall architecture of the framework is illustrated in Fig.\ref{fig2}. Given a set of training samples and their corresponding labels, the SUReID converts each image into a sequence of vectors by partitioning it into a patch grid and sending them to a teacher encoder and a student encoder, respectively. The teacher encoder adopts the ViT model that has been pretrained on the corresponding dataset with some tricks. It does not undergo back propagation during the training phase and will remain unused in the testing phase. The student encoder adopts ViT/Deit with the HTS strategy \cite{ref34}. During the training phase, the HTS strategy will record the importance of each token, and all tokens are participated in the forward propagation. Then, the output class token from teacher model will calculate the NPKD with the class token of student model. During inference, the less informative tokens are discarded, and only the rest important tokens are put forward for calculation.

\subsection{Hierarchical token sparsification strategy}\label{subsec3}

\begin{figure*}[htbp]
\centering
\includegraphics[width=0.55\textwidth]{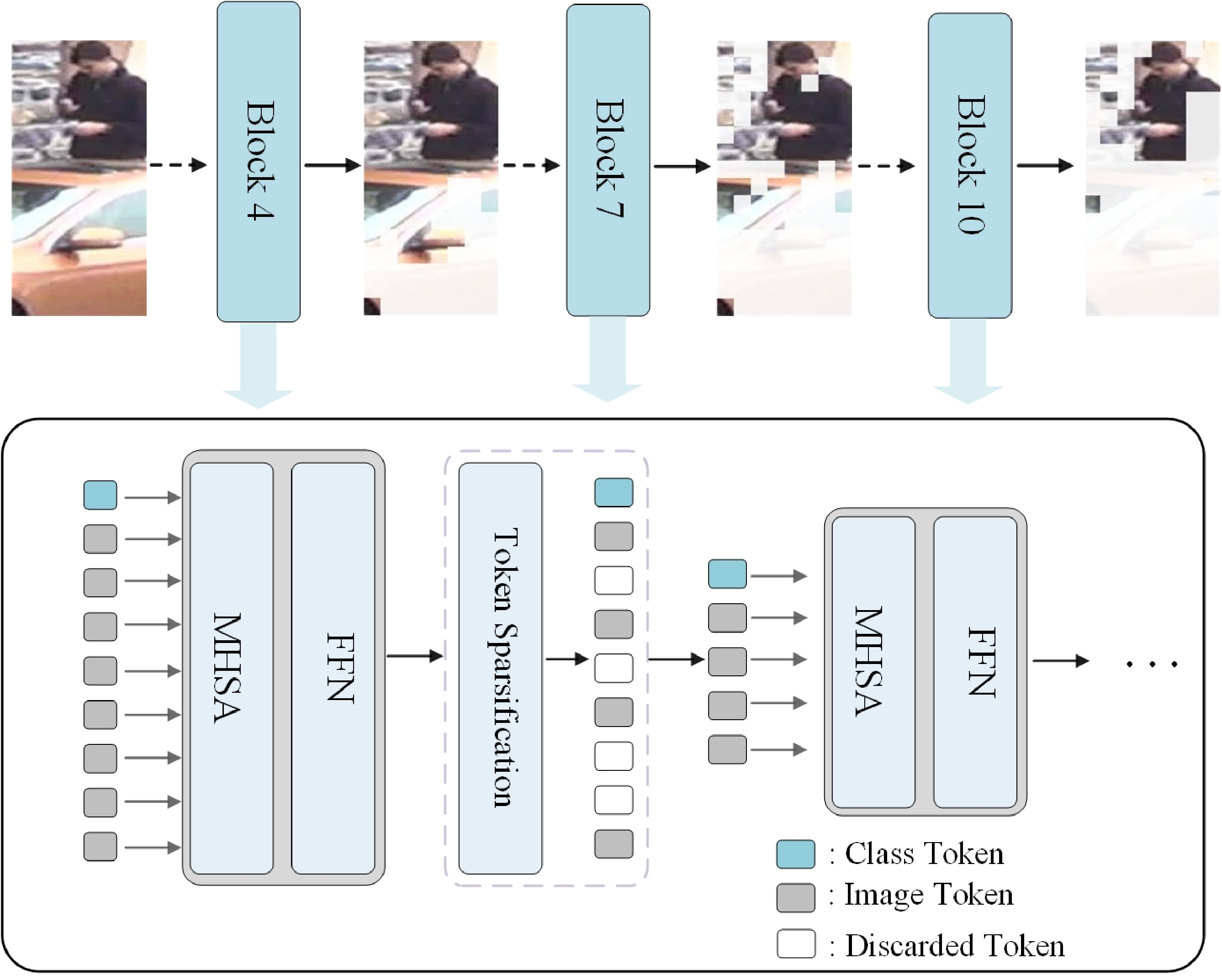}
\caption{Illustration of the HTS strategy in student encoder.}\label{fig3}
\end{figure*}

To tackle the occluded ReID problem, most previous works process the whole person image information and locate the discriminative human body parts by utilizing extra attention module or pose estimation model. This paper utilizes the HTS strategy to prune uninformative information, such as obstacles and background noise, in order to overcome the occlusion problem, as shown in Fig.\ref{fig3}.

Given an input image $x\in R^{H\times W\times d}$, transformer requires it to be divided into $N=(H\times W)/P^2$ non-overlapping patches $\{x_i\}_i^N$, where $H,W,d$ and $P$ respectively denotes the image height, width, channel and patch stride, and the image token can be denoted as $x_i \in R^{(N+1)\times C}$.

Formally, the HTS strategy works with a decision module to predict which image token should be kept. In this way, the class token is not shown and its decision is always set to 1. The binary decision mask $\hat{D} \in \{0,1\}$ is used to indicate whether to drop or keep each token. The decision mask element is initially set to 1 and subsequently updated progressively. To predict the probability of keeping or dropping the tokens, a multilayer perceptron (MLP) is utilized to reduce the image tokens dimension and obtain binary decision probabilities $\pi$ with a Softmax layer:

\begin{equation}
\pi =Softmax\left( MLP\left( {{{\hat{D}}}_{i}}\times{{x}_{i}} \right) \right)\in {{R}^{N\times2}} \label{eq.1}
\end{equation}
where $\pi_{i,0}$ means the probability of dropping the $i-th$ token and $\pi_{i,1}$ is the probability of keeping it. 

Since the binary decisions are non-differentiable, the Gumble-Softmax technique is applied to make the whole framework end-to-end trainable. The prediction probabilities $\pi$ is sampled with the Gumble-Softmax is formulated as follows:

\begin{equation}
D=Gumble-softmax{{(\pi )}_{*,~1}}\in {{\{0,1\}}^{N}}
\end{equation}
where the index "1" to make $D$ represent mask of the kept tokens. The output of Gumble-Softmax is one-hot tensor, of which the expectation equals $\pi$ exactly. With each iteration, the decision mask will be updated, and the update manner of the decision mask is shown as follows:

\begin{equation}
\hat{D}=~\hat{D} \times D
\end{equation}

Followed attention masking strategy in the DynamicViT, the tokens where $\hat{D}=0$ are not simply discarded but cut down the interactions between the pruned tokens and other tokens. The attention matrix with the masking strategy is formulated as follows:

\begin{equation}
A=\frac{Q{{K}^{T}}}{\sqrt{C}}\in {{R}^{N\times N}}
\end{equation}

\begin{equation}
{{G}_{ij}}=\left\{ \begin{matrix}
   1,~i=j  \\
   {{{\hat{D}}}_{j}},~i\ne j  \\
\end{matrix}~~1\le i,~j\le N \right.
\end{equation}

\begin{equation}
{{\tilde{A}}_{ij}}=~\frac{\exp \left( {{A}_{ij}} \right){{G}_{ij}}}{\mathop{\sum }_{k=1}^{N}\exp \left( {{A}_{ik}} \right){{G}_{ij}}},~1\le i,~j\le N
\end{equation}
where ${{G}_{ij}}=1$ means the $j-th$ token will contribute to the update of the $i-th$ token. If $\hat{D}_j=0$, the $j-th$ token will not contribute to any tokens other than itself.

Furthermore, inspired by the class token attention-based methods, a class token attention reweight module is proposed to strengthen the informative tokens. The similarity scores between the class token and other tokens are calculated to reweight the image tokens, and the class token attention $Attn_{cls}$ is formulated as follows:

\begin{equation}
Att{{n}_{cls}}=Softmax\left( \frac{{{q}_{cls}}\times{{K}^{T}}}{\sqrt{C}} \right)
\end{equation}
where $q_{cls}$ denotes the class token of query vector. In multi-head self-attention layer, the attention score is calculated with the average of all heads. Then, the class token attention is reweighted to the image tokens of the final layer. The final output $x_{output}$ is formulated as follows:

\begin{equation}
{{x}_{output}}=~x+~\lambda \times concat\left( {{x}_{class}}\cdot Att{{n}_{cls}}\times{{x}_{image}} \right)
\end{equation}

where $concat(\cdot)$ represents feature a concatenate operation and $\lambda$ is a learnable scalar initialized to 0.5. $x_{class}$ and $x_{image}$ represent class token and image tokens, respectively. Notably, the class token attention reweight module is applied only during the training phase and has no effect on inference speed.

To maintain a predefined ratio of tokens, a set of target ratios for $S$ stages, $p=[p^1,…,p^s]$ is established. The prediction module is supervised by MSE loss and the loss function $L_{ratio}$ formulated as follows:

\begin{equation}
{{L}_{ratio}}=\frac{1}{BS}\underset{b=1}{\overset{BS}{\mathop \sum }}\,\underset{j=1}{\overset{s}{\mathop \sum }}\,||{{p}^{j}}-\frac{1}{N}\underset{i=1}{\overset{N}{\mathop \sum }}\,\hat{D}_{i}^{b,j}|{{|}^{2}}
\end{equation}

\subsection{Non-parametric feature alignment knowledge distillation}\label{subsec}

After implementing the HTS strategy on the student encoder, inference time is markedly reduced as fewer tokens are computed. Nevertheless, the improved inference speed is not without cost. Discarded tokens may contain valuable human-body information, leading to decreased performance. To improve the remaining tokens' representation capabilities, the SUReID proposes NPKD strategy to guide the student encoder in learning robust feature representation.
The NPKD considers several factors which may influence the ReID model’s performance. The current mainstream works train a ReID model by jointly utilizing metric learning and representation learning methods. During the testing phase, the classifier head of the ReID model is dropped, and the similarity of feature embeddings is calculated to determine person category. Learning a strong feature representation appears to hold greater significance. In this way, the NPKD simultaneously adopts a feature-based KD and a response-based KD to enhance the feature representation capability.
However, the feature-based KD commonly suffers from the feature dimension mismatch between teacher and student. A frequently adopted strategy is to add a parametric module like a linear transformation layer. Instead, the NPKD implements a non-parametric interpolation method to align the feature dimension from teacher model to student model. Analysis and details will unfold in subsection \ref{subsec4}. The choice of teacher model is also critical in determining performance. Deit employs the convnet as a teacher model, implying that the inductive bias inherited from the convnet is more favorable for the transformer. In SUReID, the student encoder adopts a ViT architecture with token sparsification strategy. The purpose of knowledge distillation is to obtain more robust feature representation with fewer tokens. The inductive bias from the convnet may not be suitable for a transformer network using the HTS strategy. Therefore, the teacher encoder in SUReID utilizes a highly performing ViT model that shares a comparable architecture with the student model.

Given a training $x$ with one-hot label $y\in\{1,2,…,K\}$ , the output feature of the student model is denoted as $f^s\in R^C$. Followed the ReID loss setting, the feature $f^s$ is subsequently passed into the classifier to obtain the logits $g^s$ before a batch normalization (BN) layer. 

\begin{equation}
{{g}^{s}}={{W}^{s}}\times BN({{f}^{s}})
\end{equation}
where $W^s \in R^{K\times C}$ is a fully connection layer. The logits KL is defined as:

\begin{equation}
{{L}_{KL}}=KL\left( \sigma \left( {{g}^{s}} \right),~\sigma \left( {{g}^{t}} \right) \right)
\end{equation}

\begin{equation}
\sigma \left( g_{i}^{s} \right)=~\frac{\exp \left( \frac{g_{i}^{s}}{T} \right)}{\mathop{\sum }_{j=1}^{K}\exp \left( \frac{g_{j}^{s}}{T} \right)}
\end{equation}
where $KL(\cdot)$ denotes the Kullback-Leibler divergence. $g_i^s$ means the $i-th$ element of corresponding vectors and $T$ is a temperature parameter which set to 1 in this setting. $g^t$ is logits output of teacher model, which proceed similar with student model but without BN layer. 

The final NPKD loss $L_{KD}$  is formulated as a logits KL loss and a $l_2$ loss function: 

\begin{equation}
{{L}_{KD}}=~\lambda \times {{L}_{KL}}+\left| \left| {{f}^{s}}-~I\left( {{f}^{t}} \right) \right| \right|_{2}^{2}
\end{equation}
where a projector $I(\cdot)$ denotes interpolation method which is used to match the feature dimensions with no parametric costing. $\lambda$ is a hyperparameter to balance the weight of logits KL loss and feature-based KD, which is set to 0.1 in this setting. 

\begin{figure}[htbp]
\centering
\includegraphics[width=0.5\textwidth]{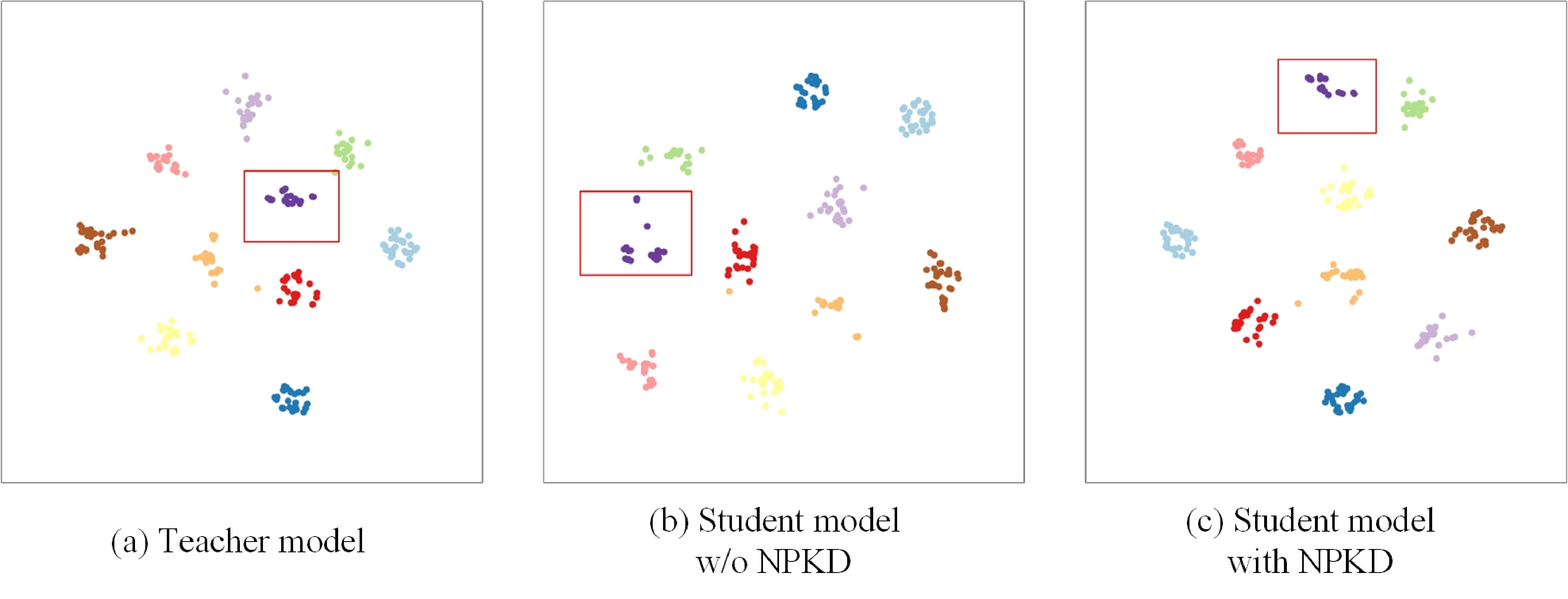}
\caption{Visualization results of test images from Occluded-DukeMTMC.}\label{fig4}
\end{figure}

With the guidance of the pre-trained teacher encoder, the distilled student encoder is compactly clustered within the same class and distinctly separated across different classes. As shown in Fig.\ref{fig4}, the red rectangle indicates that the feature embeddings with a purple color in the distilled student encoder are more compact compared to the student model without distillation. Furthermore, the feature embedding of the distilled student model exhibits a similar representation of feature embeddings to the teacher model.

The total loss for training can be formulated as:
\begin{equation}
{{L}_{SUReID}}=\alpha \times ({{L}_{KD}}+{{\lambda }_{ratio}}\times{{L}_{ratio}})+\beta \times \left( {{L}_{cls}}+{{L}_{tri}} \right) \label{eq.14}
\end{equation}
where the $\lambda_{ratio}$ is set to 2 in this setting. $L_{cls}$ and $L_{tri}$ denote the classification loss and triplet loss, which are commonly used loss function in person ReID task.
 
\subsection{Noise occlusion data augmentation}\label{subsec3}

One of the challenges in deploying pedestrian re-identification tasks lies in the wide range of occlusions present in realistic scenes. Therefore, a noise-occluded data augmentation (NODA) strategy is investigated to validate the robustness of the proposed SUReID. Recent methods focus on creating a generalized or data relevant occlusion augmentation strategy to bolster the feature representations. Random erasing is one of the most popular occluded data augmentation (ODA) strategies, it randomly erases a rectangular area and sets the pixel values within that area to zero or random values \cite{ref35}. The erased area is used to simulate real occlusion and make the model robust to occlusions. However, the unified occlusion shape may fail the model handle the diversified occlusions. The other ODA strategies utilize the specific occlusion information to make augmentation. FED and DPL borrow occlusion samples from the training dataset to create more occluded person images, which can be called data-related augmentation \cite{ref36}. The borrowed occlusion samples are same to the occlusion samples in the testing data, which will reduce the data variance between the training data and testing data since those image data are captured from the same place.

Different from the above two kinds of data augmentation strategies, the proposed NODA strategy utilizes the real-world occlusion to make augmentation. To meet this requirement, it is common to randomly capture occluded samples, such as garbage cans, bikes, and trees, from real-world scenes to create occlusion patches. These samples can be used to enhance the realism of the occlusion in the simulation. Actually, these occluded samples are unrelated to the training data, can be considered as noise data augmentation and have the potential to contaminate the learned features. Owing to the mechanism of the proposed SUReID, the noise information can be greatly mitigated by the HTS strategy and occlusions further improve the capability of the HTS strategy to disentangle the representations of target person. 

\begin{figure}[htbp]
\centering
\includegraphics[width=0.5\textwidth]{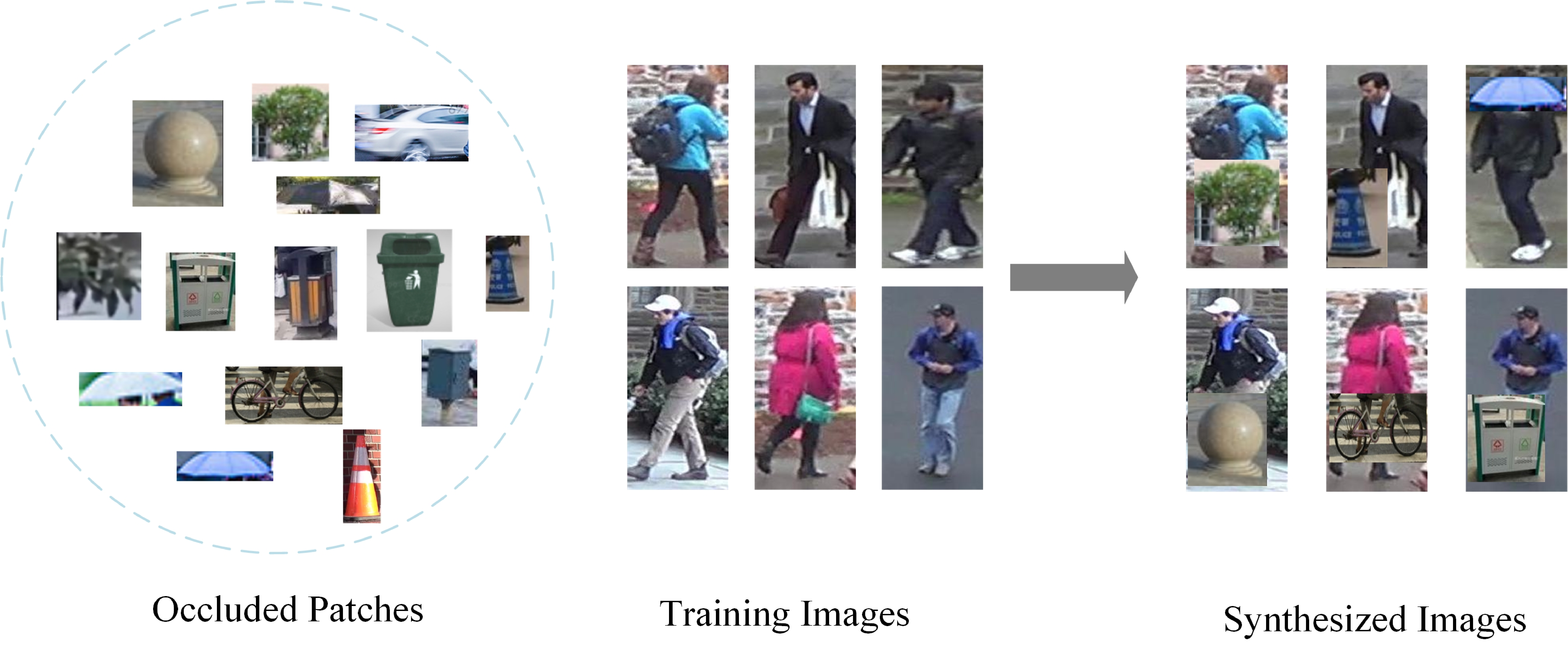}
\caption{Illustration of the augmented pedestrians by the NODA.}\label{fig5}
\end{figure}

Specifically, the NODA strategy is described as follows. Firstly, the given input images are duplicated to create two batch images. The first batch images are augmented with common augmentations, such as resize, padding, random crop, and generalized occluded data augmentation random erasing and random patch. Another batch images are also augmented with common augmentations, then applying the NODA strategy, as shown in Fig.\ref{fig5}. Secondly, occlusions empirically happen at four locations (top, bottom, left, right) approximately one-third to half of the areas. The patch $p \in R^{(3\times p_h\times p_w)}$ from the occlusion set is randomly selected to make data augmentation, where $p_h$ and $p_w$ denote the height and width. Followed the occlude strategy of FED, the pasted patch is first calculated the aspect ratio: $\alpha$ $=p_h/p_w$. When $\alpha$ is larger than 2, it implies the patch is more like a vertical occlusion, otherwise horizontal occlusion. Common augmentation, such as color jitter, random horizontal flip and random crop, are also applied on the patch for increasing its varieties. Finally, the patches are resized to $R^{((H/3~H/2,W))}$ and $R^{((H,W/3~W/2))}$ according to the occlusion type (horizontal or vertical), respectively. The augmented patches are then randomly pasted onto input image.

\section{Experiments}\label{sec4}
This section conducts the comparison experiment to verify the validity of the SUReID framework. A set of ablation studies are reported to validate the effectiveness of each component. Last, some visualization results are shown to further demonstrate the proposed method can focus on discriminative human parts for the occluded ReID task.

\subsection{Datasets and Evaluation Metrics}\label{subsec4}
 
The experiment is evaluated on five challenging person ReID datasets to verify the effectiveness of the proposed SUReID. The datasets include the occluded pedestrian datasets Occluded-DukeMTMC, Occluded-REID and Partial-REID, the holistic pedestrian datasets DukeMTMC-reID and Market-1501 \cite{ref37,ref38,ref39, ref40}.

\textbf{Occluded-DukeMTMC}:This dataset is derived from the DukeMTMC-reID and specifically designed for occluded person re-ID. The samples in this dataset are collected from eight non-overlapping cameras. The training set consists of 15,618 images of 702 pedestrians. The testing set consists of 19,871 images of an additional 519 pedestrians, and each image contains occluded objects.

\textbf{Occluded-REID}: This dataset is specifically designed for occluded person ReID and is captured using mobile cameras. It comprises 2000 images belonging to 200 different identities. Each identity includes five full-body person images and five occluded person images with varying viewpoints and severe occlusions of different types.

\textbf{Partial-REID}: This dataset is a specially designed ReID dataset that consists of the occluded, partial and holistic pedestrian images. It comprises 600 images of 60 pedestrians. The partial images are set to query and the holistic images are set to gallery.

\textbf{Market-1501}: This dataset comprises 32,668 images of 1,501 pedestrians captured by six non-overlapping cameras. The training set consists of 12,936 images of 751 pedestrians, while the testing set contains 19,732 images of 750 pedestrians.

\textbf{DukeMTMC-reID}: This dataset comprises 36,411 images of 1,404 pedestrians captured by eight non-overlapping cameras. The training set consists of 16,522 images of 702 pedestrians, while the testing set contains 19,889 images of the remaining 702 pedestrians.

\textbf{Evaluation Metrics}: The evaluation metrics employed in this study follow the standard practices commonly used in person ReID research, namely Cumulative Matching Characteristic (CMC) curves and mean average precision (mAP). These metrics are used to assess the performance of various ReID models. All experiments are conducted in the single query mode, without utilizing re-ranking techniques to further refine the matching results.

\subsection{Implementation Details}\label{subsec4}

The teacher model adopts ViT pretrained on corresponding datasets with camera-constrained triplet loss and occluded data augmentation strategy in SRFR \cite{ref41}. The student model utilizes ViT with the HTS strategy for the Occluded-REID and Partial-REID datasets, while employing Deit-S with the HTS strategy for the other datasets. Deit is pretrained on ImageNet-1K with distillation token and ViT is pretrained on ImageNet-21K and finetuned on ImageNet-1K as the student encoder. The sparsification stages are set to the 4$th$, 7$th$, and 10$th$ layers in transformer architecture and the token keeping ratios are set to $p$, $p^2$, $p^3$, where $p$ ranges from (0, 1). The hyperparameters $\alpha$ and $\beta$ in Eq.\ref{eq.14} are configured differently for the Occluded-REID and Partial-REID datasets, with values of 2 and 1 respectively. For other datasets, the values are set to 1 for both $\alpha$ and $\beta$. All images are resized to 256 $\times$ 128 and augmented with NODA strategy. Extra color jitter is adopted on occluded-REID and partial-REID to avoid domain variance. The batch size is set to 64 with 4 images per person. During the training phase, all models are jointly trained for 200 epochs, the teacher model does not proceed back propagation. SGD optimizer is employed with a momentum of 0.9. The learning rate is initialized as 0.008 with cosine learning rate decay. All the experiments are performed with one Nvidia RTX 2080Ti GPU using the PyTorch toolbox.

\subsection{Comparison Experiments}\label{subsec4}

This subsection will compare the proposed SUReID with several methods on both occluded ReID datasets and holistic ReID datasets. The Rank-1 accuracy, mAP and throughput under different keeping ratios are reported to demonstrated the effectiveness of the proposed SUReID. The throughput is measured on a single NVIDIA RTX 2080Ti GPU with batch size fixed to 32.

\noindent\textbf{Results on Occluded-DukeMTMC.} To verify the performance and effectiveness of the proposed SUReID, it is applied to the commonly used occlusion dataset Occluded-DukeMTMC. The SUReID is compared with the state-of-the-art methods that provide the official code, which facilitates the testing of the model's inference speed, as shown in Table.\ref{tab1}. The PFD, PGFA, POS, PVPM, HOReID and ISP methods are all utilizing an extra pose estimation model or human parsing to supervise the backbone network focus the human body features \cite{ref42}. The PFD adopts HR-Net as the pose estimation model based on a ViT backbone, which achieves great performance. However, the ViT backbone and HR-Net are both weighty models, it dramatically slows the inference speed \cite{ref43}. The POS and HOReID also utilize HR-Net as the pose estimation model, but they use ResNet50 as their backbone by removing the last down-sampling operation \cite{ref44}. With their specially designed module for handling occlusions, the POS and HOReID methods achieve satisfactory performance with relatively slower inference speed. The proposed SUReID (Deit) model achieves almost tenfold increase in inference speed compared to the above methods while exhibiting higher performance. The PGFA and PVPM methods also leverage the pose estimation model, however, the estimation model is not trained in an end-to-end manner. The pose heatmaps are generated using Openpose before training, thus the extra pose estimation model will not impact the inference time. Nevertheless, the PGFA and PVPM are based on part-level feature methods, which requires larger size image, leading to increased computational costs. The ISP relies on a human semantic parsing method to locate both human body parts and personal belongings at pixel-level. The pseudo-labels for discriminative foreground parts are generated separately, it has no effect on the testing time. However, the ISP needs high-resolution representations for containing more semantic information. Therefore, the ISP adopts HR-Net as the backbone network for a larger feature map, which is a tradeoff between the performance and inference time. Compared to the ISP, the SUReID achieves the higher performance but with nearly triple the speed. The other set of methods are not assisted by extra auxiliary model but utilize a transformer architecture. The PAT uses ResNet50 as the backbone network and a transformer encoder-decoder architecture to disentangle the discriminative human body parts. Since the official code for PAT is not available, we simulate the PAT framework using a ResNet50 backbone and two self-attention modules to measure its inference speed. Even though the experiment results show the PAT achieves impressive result, the proposed SUReID still surpasses it for both performance and inference speed by a large margin. The TransReID and FED methods are pure transformer networks which show great potential on occluded ReID task. Nonetheless, transformer is a bulky network which is not suitable for real-time applications. The "*" means the Transformer network is in a sliding-window setting.
 The proposed SUReID is also constructed on transformer network, but the inference model is more light-weight model Deit. Besides, different from the methods dealing with the whole image information, the proposed SUReID only calculates the discriminative human-body feature for robust feature representation, achieving performing performance and higher inference speed. 

\begin{table}[h]
\caption{Performance comparison on Occluded-DukeMTMC.}\label{tab1}
\begin{center}
\begin{tabular}{@{}lllll@{}}
\toprule
Model                  & \multicolumn{1}{c}{Size}    & Throughput(img/s) & Rank-1 & mAP  \\ \midrule
PFD                    & 256$\times$128 & \multicolumn{1}{c}{119} & 67.7   & 60.1 \\ 
PFD$^*$                    & 256$\times$128 & \multicolumn{1}{c}{95} & 69.5   & \textbf{61.8} \\
PGFA                   & 384$\times$128 & \multicolumn{1}{c}{744}       & 51.4   & 37.3 \\
POS                    & 256$\times$128 & \multicolumn{1}{c}{137}        & 65.0   & 54.0 \\
PVPM                   & 384$\times$128 & \multicolumn{1}{c}{630 }         & 47.0   & 37.7 \\
HOReID                 & 256$\times$128 & \multicolumn{1}{c}{141}          & 55.1   & 43.8 \\ 
ISP                    & 256$\times$128 & \multicolumn{1}{c}{554}           & 62.8   & 52.3 \\
TransReID              & 256$\times$128 & \multicolumn{1}{c}{379}            & 64.2   & 55.7 \\
TransReID$^*$          & 256$\times$128 & \multicolumn{1}{c}{169}            & 66.4   & 59.2 \\
PAT                    & 256$\times$128 & \multicolumn{1}{c}{849}             & 64.5   & 53.6 \\
FED                    & 256$\times$128 & \multicolumn{1}{c}{411}              & 68.1  & 56.4 \\
SRFR                   & 256$\times$128 & \multicolumn{1}{c}{55}              & \textbf{71.4}   & 60.6 \\
SUReID            & 256$\times$128 & \multicolumn{1}{c}{\textbf{1585}}              & 65.8   & 54.9 \\
\bottomrule
\end{tabular}
\end{center}
\end{table}

\noindent\textbf{Results on Occluded-REID and Partial-REID} To further evaluate the proposed SUReID, the experiments on Occluded-REID and Partial-REID are conducted to compare the results with other methods. Since the two datasets are too small, the Market-1501 training set is used to train the model. Therefore, it can be viewed as a cross-domain setting. In the two datasets, the student encoder is set to ViT with HTS strategy for 0.7 ratio. The Deit is a light-weight model which is easy to overfit on the two small datasets. The SUReID achieves outstanding performance on the Occluded-REID dataset, with Rank-1 accuracy of 86.8\% and mAP of 80.7\%, nearly reaching the best performed method, which is shown in Table.\ref{tab2}. The outstanding performance demonstrates the scalability of the SUReID since it can suppress interference from the occluded objects. The SUReID also performs well on Partial-REID, although it achieves slightly lower accuracy compared to PAT, with 2.7\% decrease in Rank-1 accuracy and 1.6\% decrease in Rank-3 accuracy. This is due to the fact that the Partial-REID contains partial person images rather than occluded samples. The HTS strategy prunes the tokens that are relate to the human information. 

\begin{table*}[h]
\begin{center}
\begin{minipage}{\textwidth}
\caption{Performance comparison on Occluded-REID and Partial-REID datasets.}\label{tab2}
\begin{tabular*}{\textwidth}{@{\extracolsep{\fill}}lcccccc@{\extracolsep{\fill}}}
\toprule%
& \multicolumn{2}{@{}c@{}}{Occluded-REID} & \multicolumn{2}{@{}c@{}}{Partial-REID} \\
\cmidrule{2-3}\cmidrule{4-5}%
Methods & Rank-1 & mAP & Rank-1 & Rank-3 \\
\midrule
PGFA & -  &  -  & 68.0 & 80.0\\
PVPM & 70.4  &  61.2  & 78.3 & 87.7\\
HOReID &  80.3  &  70.2 & 85.3 & 91.0\\
POS &  -  &  - & 86.1 & 91.3\\
PAT &  81.6  &  72.1 & \textbf{88.0} & \textbf{92.3}\\
FED & \textbf{87.0}  &  79.4 & 84.6  & -\\
PFD & 79.8  &  81.3 & -  & -\\
PFD$^*$ & 81.5  &  \textbf{83.0} & -  & -\\
SRFR & 86.8 & 82.0 & 86.0 & 92.0 \\
SUReID & 86.8 & 80.7 & 85.3 & 90.7 \\
\toprule
\end{tabular*}
\end{minipage}
\end{center}
\end{table*}

\begin{figure*}[htbp]
\centering
\includegraphics[width=1.0\textwidth]{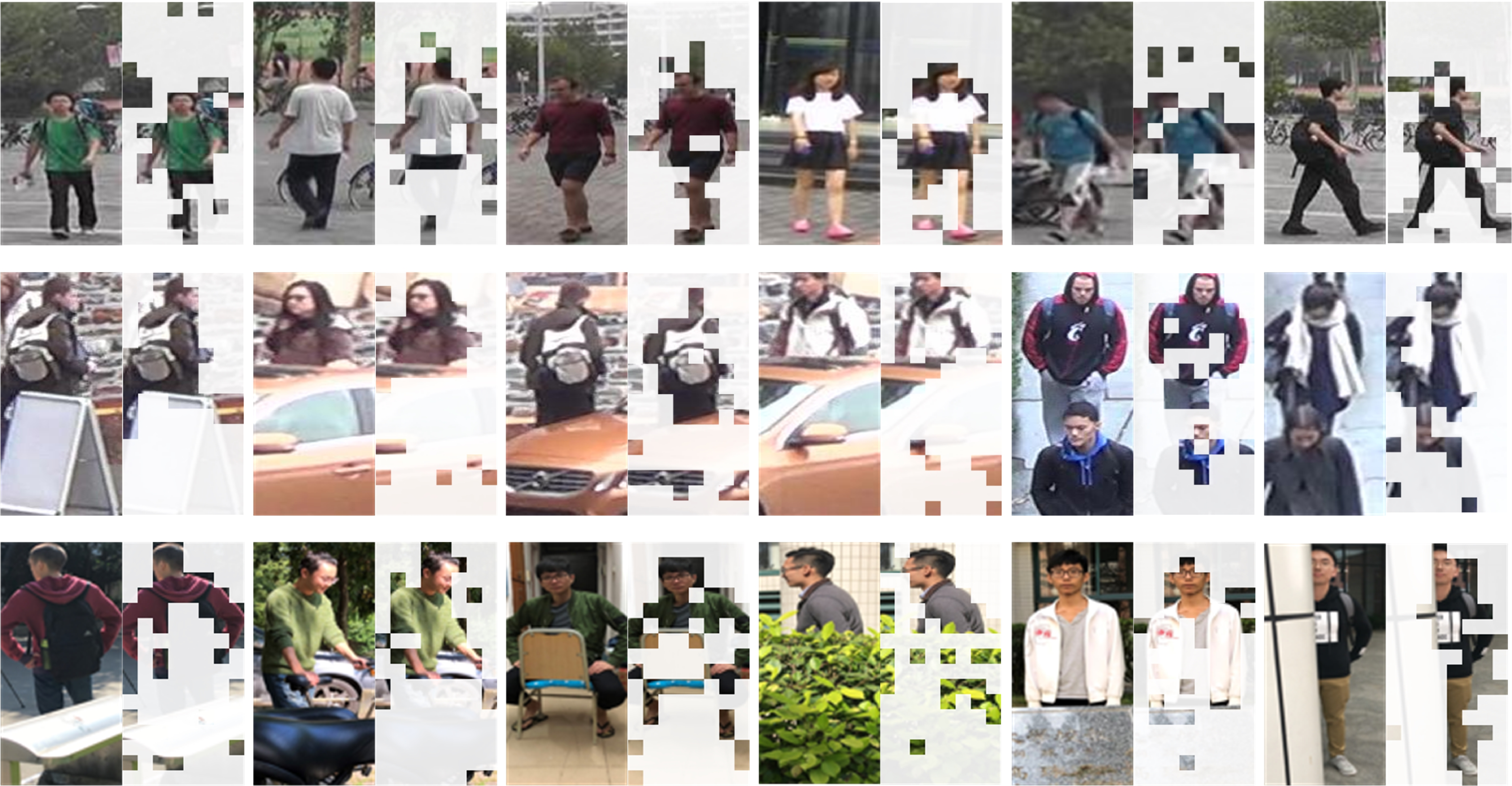}
\caption{Visualization of the sparsified tokens.}\label{fig6}
\end{figure*}

\noindent\textbf{Results on Holistic ReID} Although the SUReID is proposed to solve occlusion problem, it works well on the holistic person ReID task. In the following experiments, two challenging holistic datasets Market-1501 and DukeMTMC-reID are used to verify the performance of the proposed SUReID. The inference speed is same in Table.\ref{tab3}, which is not shown in this table for simplicity. For Market-1501 and DukeMTMC-reID, the proposed SUReID achieves the 94.5\% and 89.1\% recognition rates on Rank-1, and 87.2\% and 79.7\% accuracy on mAP. The performance of the proposed SUReID is slightly lower than the best performing method PFD, but the inference speed of SUReID is more than ten times faster than it. The reason for the superior performance can be summarized as follows. Firstly, the holistic person images contain background, which can be thought noise information. The SUReID can effectively reduce the interference from background noise by the HTS strategy, the illustrations are shown in Fig.\ref{fig6}. Secondly, the NPKD can improve feature representation capability of the inference model by distilling the prior knowledge from teacher model. 

\begin{table*}[h]
\begin{center}
\begin{minipage}{\textwidth}
\caption{Performance comparison on Market-1501 and DukeMTMC-reID datasets.}\label{tab3}
\begin{tabular*}{\textwidth}{@{\extracolsep{\fill}}lcccccc@{\extracolsep{\fill}}}
\toprule%
& \multicolumn{2}{@{}c@{}}{Market-1501} & \multicolumn{2}{@{}c@{}}{DukeMTMC-reID} \\
\cmidrule{2-3}\cmidrule{4-5}%
Methods & Rank-1 & mAP & Rank-1 & mAP \\
\midrule
PGFA  & 91.2 & 76.8  & 82.6 & 65.5\\
HOReID  & 94.2 & 84.9  & 86.9 & 75.6\\
ISP & 95.3  &  88.6  & 89.6 & 80.0\\
POS  & 95.0  &  86.2  & 88.7 & 76.7\\
TransReID  & 95.2 & 88.9 & 89.6 & 80.6\\
TransReID$^*$  & 95.0 & 88.2 & 90.7 & 82.0\\
PAT  & 95.4  &  88.0  & 88.8 & 78.2\\
FED  & 95.0  &  86.3  & 89.4 & 78.0\\
PFD  & 95.5  &  89.6 & 90.6 & 82.2\\
PFD$^*$  & 95.5  &  89.7 & \textbf{91.2} & \textbf{83.2}\\
SRFR & \textbf{95.9}  &  \textbf{90.2} & 90.9 & 82.0\\
SUReID &  94.5  &  87.2 & 89.1 & 79.7\\
\toprule
\end{tabular*}
\end{minipage}
\end{center}
\end{table*}

\subsection{Ablation Studies}\label{subsec4}
 
In this subsection, the ablation experiments of the HTS strategy, NPKD and NODA are separately conducted to validate the effectiveness of each component. The experimental results are shown in Table.\ref{tab4}. Index-1 shows the performance of vanilla Deit on Occluded-DukeMTMC, which shows it achieves 57.6\% Rank-1 accuracy and 49.2\% mAP. Index-2 demonstrates that the performance degrades when using the HTS strategy, but it has the advantage of lower computational complexity. Then, the NPKD is adopted to enhance the feature representation capability, while maintaining the same throughput. Index-3 indicates a substantial improvement on Rank-1 accuracy and mAP, and Fig.4 illustrates the feature embedding distribution is more similar to the teacher model, which indicates the effectiveness of the proposed NPKD. Finally, in index-4 shows the SUReID achieves the best accuracy at 65.8\% Rank-1 and 54.9\% mAP, demonstrating its robustness for occlusion samples in real-life scenarios.

\begin{table}[h]
\begin{center}
\caption{Performance comparison with different components}\label{tab4}
\begin{tabular}{@{}llllllll@{}}
\toprule
Index & HTS & NPKD & NODA & Throughput(img/s)& Rank-1 & mAP\\
\midrule
1   &  &  &  & \multicolumn{1}{c}{1074} & 57.6 & 49.2 \\
2   & \hspace{2mm}\checkmark  &   &  & \multicolumn{1}{c}{1585} & 55.6 & 46.7 \\
3   &  \hspace{2mm}\checkmark& \hspace{3mm}\checkmark  &  & \multicolumn{1}{c}{1585} &64.8 & 53.2 \\
4   & \hspace{2mm}\checkmark & \hspace{3mm}\checkmark & \hspace{3mm}\checkmark & \multicolumn{1}{c}{1585} & 65.8 & 54.9 \\
\toprule
\end{tabular}
\end{center}
\end{table}

The token keeping ratio in HTS strategy also effects the ReID model performance and inference speed. To obtain a better complexity/accuracy tradeoff, the SUReID is trained with different keeping ratio value $p$. Table.\ref{tab5} shows the performance and throughput of the SUReID on Occluded-DukeMTMC with different token keeping ratios. Intuitively, when $p$ is small, the inference speed will be accelerated but may discard informative information, resulting in a relatively poor performance. However, thanks to the distillation mechanism of the proposed SUReID framework, even with a relatively smaller value of $p$, it achieves remarkably higher speed while maintaining satisfactory performance. As the token keeping ratio $p$ increases, the inference speed decreases while the performance improves. When $p$ is equal to 0.7, the SUReID achieves the optimal balance between complexity and accuracy, making it the most favorable tradeoff.

\begin{table}[h]
\caption{The effect of the sparsification ratio}\label{tab5}
\begin{center}
\begin{tabular}{@{}llll@{}}
\toprule
\multirow{2}{*}{Sparse ratio} & \multicolumn{2}{l}{Occluded-DukeMTMC} &  \\ \cmidrule{2-4}
                     & Throughput(img/s)   & Rank-1           & mAP   \\ \midrule
$p$=0.5    & \multicolumn{1}{c}{2269}   & 64.2      & 53.2  \\
$p$=0.6    & \multicolumn{1}{c}{1843}   & 65.6      & 54.2  \\
$p$=0.7    & \multicolumn{1}{c}{1585}   & 65.8      & 54.9   \\
$p$=0.8    & \multicolumn{1}{c}{1392}   & 65.5      & 55.3    \\
$p$=0.9    & \multicolumn{1}{c}{1279}   & 65.9      & 55.7    \\
\toprule
\end{tabular}
\end{center}
\end{table}

Normally, the feature dimensions of teacher model are larger than student model, which result in feature misalignment issue. In this subsection, the ablation studies are conducted to investigate the effectiveness of the two feature dimension aligning methods for the SUReID. "T-S" means align feature dimension from the teacher model to student model; "S-T" denotes align feature dimension from the student model to teacher model. For parametric method, a Linear projection layer is adopted to align the feature dimension. It can be seen from the Table.\ref{tab6}, the performance degrades when using the Linear layer to align feature dimension. It can be concluded that the learnable parameters produced by parametric method will contaminate the learned feature embedding. The non-parametric method interpolation shows better performance compared to the parametric method. It can be seen that the interpolation method works effectively by compressing the features with minimal information loss. Furthermore, the performance of "T-S" is better than "S-T", it because the feature alignment from student model to teacher model increase irrelevant feature information since teachers typically have larger feature dimensions.

\begin{table}[h]
\caption{Feature alignment method}\label{tab6}
\begin{center}
\begin{tabular}{cccll}
\toprule
\multirow{2}{*}{Align method} & \multicolumn{2}{c}{Occluded-DukeMTMC} &  &  \\ \cmidrule{2-4}
                        & Rank-1           & mAP            &  &  \\ \midrule
S-T(Linear)             & 56.6             & 46.7           &  &  \\
T-S(Linear)             & 57.2             & 47.1           &  &  \\
S-T(Interpolation)      & 64.3             & 53.3           &  &  \\
T-S(Interpolation)      & 65.8             & 54.9           &  & \\
\toprule
\end{tabular}
\end{center}
\end{table}

\begin{table}[]
\caption{The effectiveness of the NODA}\label{tab7}
\begin{center}
\begin{tabular}{@{}lllll@{}}
\toprule
\multirow{2}{*}{Method} & \multicolumn{2}{l}{Occluded-Duke} &  &  \\ \cmidrule{2-4}
                        & Rank-1           & mAP            &  &  \\ \midrule
Vanilla ViT             & 61.9             & 54.0           &  &  \\
Vanilla ViT+ NODA       & 61.1             & 51.6           &  &  \\
SUViT w/o NODA          & 64.8             & 53.2           &  &  \\
SUReID                   & 65.8             & 54.9           &  &  \\
\toprule
\end{tabular}
\end{center}
\end{table}

Table.\ref{tab7} shows the effectiveness of the NODA. The first line shows the performance of vanilla ViT achieves 61.9\% and 54.0\% accuracy on Rank-1/mAP. When the vanilla ViT is augmented by the NODA, the performance of Rank-1/mAP decreases 0.8\%/2.4\%. The NODA is constructed by the common occlusions in real-life scenarios, which is a data irrelevant augmentation. When the NODA is applied to vanilla ViT, the noise information will contaminate the learned feature representation. However, the SUReID obtains better result when it adopts the NODA strategy. The main reason for the improving result can be concluded as follows. The SUReID adopts HTS strategy to prune the less informative tokens, the capability of disentangling the representation of body part can be further improved when introducing the occluded samples. For both considering the inference speed and robustness to data irrelevant occlusions, the SUReID is demonstrated as a more efficient and suitable ReID framework for deployment.

\subsection{Visualization}\label{subsec6}

Fig.\ref{fig6} presents visualization results that demonstrate the interpretability of the SUReID framework. The images on the left are original person images, while those on the right are person images produced using the HTS strategy. These masked patches indicate the corresponding discarded tokens. As the figure illustrated, the background noise, occluded samples and irrelevant pedestrians are treated as inattentive information which are discarded. This demonstrates that the proposed SUReID can focus on the discriminative human body part features. Additionally, the inference model solely calculates the retained tokens, resulting in a fast and effective ReID framework.

\section{Conclusion}\label{sec5}
This paper presents SUReID, a lightweight, efficient and robust framework for occluded person ReID task. Unlike other methods which depend on visibility cues from outside tools, the SUReID provides a new solution to tackle occlusion problem. Based on the HTS strategy, the uninformative information is discarded, significantly reducing the occlusion interference and inference time. The NPKD strategy is proposed to improve the feature representation capability of the rest tokens. Moreover, the NODA is proposed to validate the robustness of the SUReID for unseen occlusion samples. The occluded samples can further improve the capability of the HTS strategy for disentangling the feature representation. Jointly optimizing the above explorations, extensive experiments on five popular datasets demonstrate the effectiveness of SUReID. The proposed SUReID framework deals with occlusion issues while improving inference speed, making it a favourable option for ReID applications.

\section{Limitation and future works}\label{sec6}
The findings of this study have to be seen in light of some limitations. The kept token ratios in SUReID are predefined values, resulting in a set number of tokens being discarded for all person images. However, the proportion of informative information in person images varies. Thus, discarding the same number of tokens for all person images may not be suitable. In future work, we will focus on devising a new strategy to adaptively discard uninformative tokens and explore a better trade-off between speed and accuracy. Moreover, the performance of SUReID heavily relies on the performance of the teacher model. We believe that our method has the potential for further improvement by exploring more robust teacher models.

\vfill

\end{document}